\documentclass[letterpaper, 10 pt, conference]{ieeeconf}  

\IEEEoverridecommandlockouts                              

\usepackage{mathptmx}       
\usepackage{helvet}         
\usepackage{courier}        
\usepackage{graphicx}        
\usepackage{multicol}        
\usepackage{multirow}

\usepackage[english]{babel}

\usepackage{amsmath}
\usepackage{amsfonts}
\usepackage{graphicx}
\usepackage{relsize}
\usepackage{amssymb}  
\usepackage{algorithm}
\usepackage{algpseudocode}
\usepackage{listings}
\usepackage{longtable}
\usepackage{textcomp}
\usepackage{url}
\usepackage{alltt}
\usepackage{caption}
\usepackage{subfig}
\usepackage{multicol}
\usepackage{color}
\usepackage{makecell}
\usepackage{tabularx}

\newcommand{\ws}[1]{{#1}}

\title{
Evaluating Methods for End-User Creation of Robot Task Plans
}

\author{Chris Paxton,$^{1}$ Felix Jonathan,$^{1}$ Andrew Hundt,$^{1}$ Bilge Mutlu,$^{2}$ and Gregory D. Hager$^{1}$
\thanks{$^{1}$Department of Computer Science, Johns Hopkins University, 3400 N Charles Street, Baltimore, MD 21218, USA, {\tt\small\{cpaxton, fjonath1, ahundt1, hager\}@jhu.edu}}
\thanks{$^{2}$Department of Computer Sciences, University of Wisconsin--Madison, 1210 W Dayton Street, Madison, WI 53706, USA {\tt\small bilge@cs.wisc.edu}}}

\begin{document}

\maketitle
\thispagestyle{empty}
\pagestyle{empty}

\newcommand{\myabstract}[0]{
How can we enable users to create effective, perception-driven task plans for collaborative robots?
We conducted a 35-person user study with the Behavior Tree-based CoSTAR system to determine which strategies for end user creation of generalizable robot task plans are most usable and effective.
CoSTAR allows domain experts to author complex, perceptually grounded task plans for collaborative robots.
As a part of CoSTAR's wide range of capabilities, it allows users to specify SmartMoves: abstract goals such as ``pick up component A from the right side of the table.''
Users were asked to perform pick-and-place assembly tasks with either SmartMoves or one of three simpler baseline versions of CoSTAR.
Overall, participants found CoSTAR to be highly usable, with an average System Usability Scale score of 73.4 out of 100.
SmartMove also helped users perform tasks faster and more effectively; all
SmartMove users completed the first two tasks, while not all users completed the
tasks using the other strategies. SmartMove users showed better performance for incorporating perception across all three tasks.}

\begin{abstract}
  \myabstract
\end{abstract}

\section{Introduction}

The relatively recent development of human-safe robots has spurred a new wave of
innovation in commercial, end-user-programmable systems such as the
Rethink Robotics Sawyer, Universal Robots UR5, and Franka Emika
robots.
Robotic systems in the laboratory are becoming ever more intelligent, flexible, and robust, and these advances are increasingly translated into applications in industry.
As a result, today’s manufacturers seek skilled workers with strong
problem-solving skills and a STEM background as a part of the transformation to
``Industry 4.0,'' focused on smart and interconnected
facilities~\cite{giffi2015skills}.
The workers and factories of the future require intelligent, powerful tools that
allow them to utilize the best capabilities of modern robots.
However, whether or not even skilled workers and expert users can translate the
complex perception, planning, and control capabilities offered by modern
collaborative robots into industrial efficiency is unknown.

These developments have led to a growing interest in making it easy for domain
experts to transfer knowledge to collaborative robots, either through a user
interface~\cite{nguyen2013ros,mateo2014hammer,brunner2016rafcon,paxton2017costar},
natural language~\cite{misra2014tell}, or learning from
demonstration~\cite{ahmadzadeh2015learning,levine2015learning,dianov2016extracting,paxton2016want}.
To take full advantage of these systems, the human user must have an accurate
mental model of a robot's capabilities~\cite{sharp2007interaction}.
Therein lies a potential problem, as people tend to think of robot motions
abstractly, e.g., ``grab the next available component for this assembly,''
instead of in terms of basic robotics concepts like fixed spatial positions.

\begin{figure}[bt]
\centering
\includegraphics[width=\columnwidth]{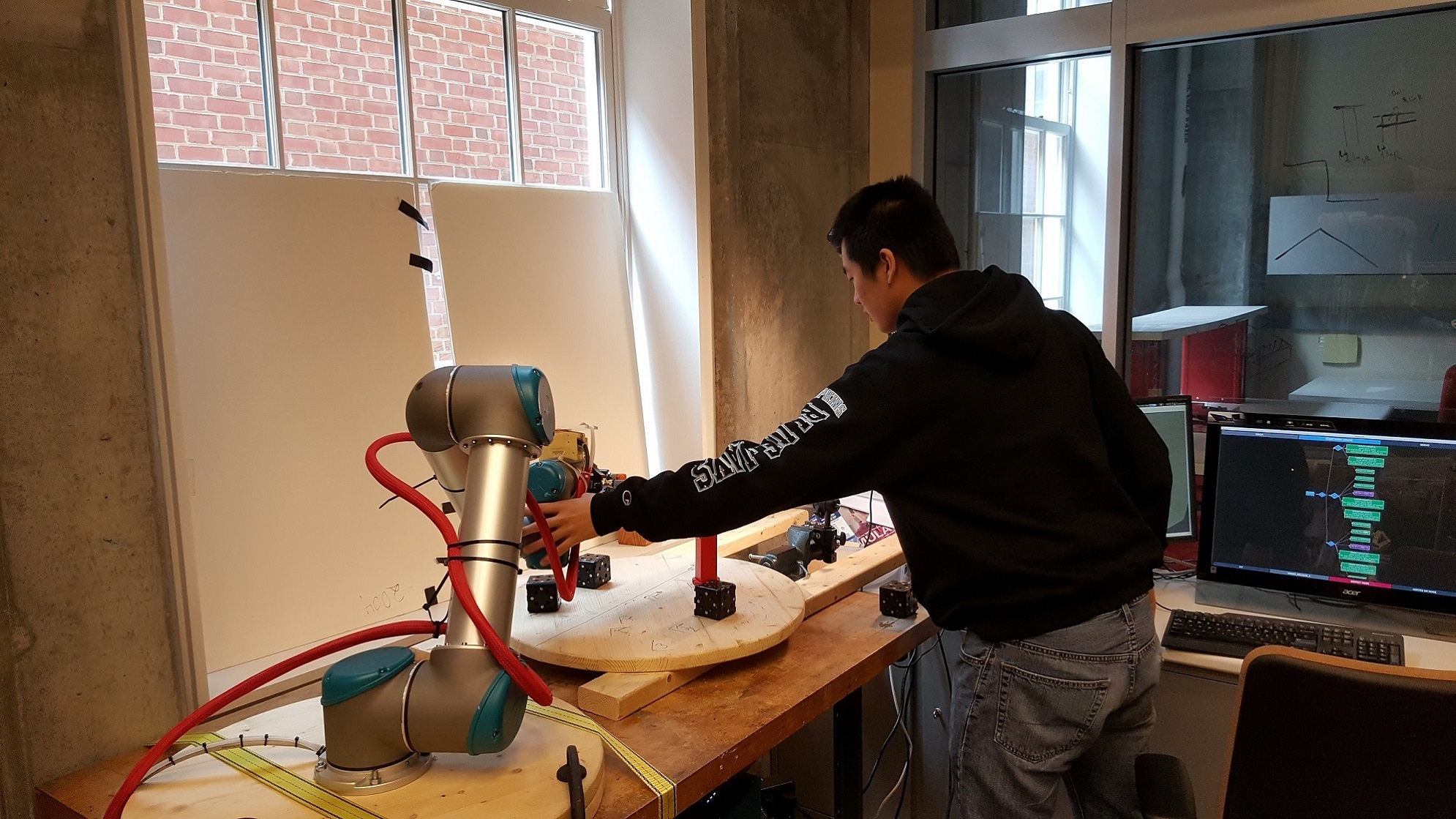}
\vskip 0.1cm
\includegraphics[width=\columnwidth]{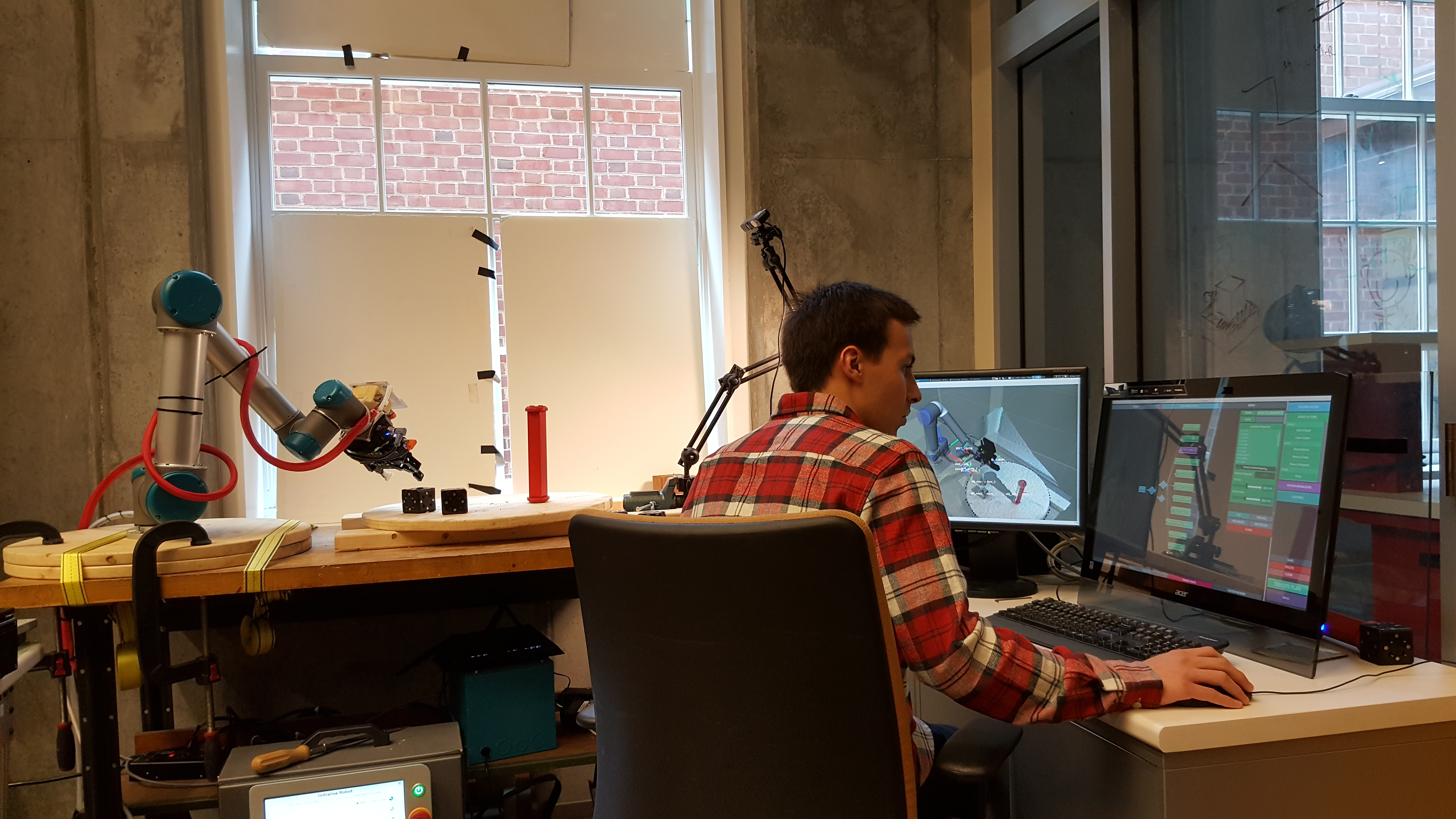}
\caption{Users interacted with the CoSTAR system in different ways. Left: a
  user positions the robot to teach it a new waypoint. Right: a user fine tunes
part of a tree using the BT-based user interface.}
\label{fig:users}
\vskip -0.5cm
\end{figure}

We previously proposed the CoSTAR system as a potential solution.
\footnote[3]{Source code for the CoSTAR system is available on GitHub: \url{https://github.com/cpaxton/costar_stack}}
CoSTAR offers a set of planning and perceptual capabilities, united through a
Behavior Tree-based task plan editor, that is designed to enable expert users to
author complex plans for collaborative robots.  We also demonstrated that
CoSTAR allows users to solve a wide variety of problems such as sanding,
assembly, and wire bending~\cite{paxton2017costar}.
It provides users a great deal of flexibility by offering many different ways
to specify actions.

These new ``SmartMoves'' were designed mimic the abstract way people think about manipulation
tasks. To perform a \texttt{SmartGrasp} or a \texttt{SmartRelease} operation,
one of the two new actions we describe in this paper, a user demonstrates a
grasp or a position. Then they specify the associated object types and abstract
descriptions like ``left of the robot'' that they want for that operation.

In this paper, we examine the usability of CoSTAR's Behavior Tree-based user
interface, and explore different strategies for using perception to create
grounded task plans.
To represent different strategies for incorporating advanced capabilities such
as perception and motion planning, we compare three versions of the CoSTAR
system without high-level abstraction against CoSTAR with SmartMoves:
\begin{itemize}
\item[(1)] \textit{Simple}: a blind
version of the system with only the ability to servo to pre-programmed waypoints in joint space.
\item[(2)] \textit{Motion
  Planning}: a system that uses perception with motion planning to
avoid obstacles.
\item[(3)] \textit{Relative Motion}: a system that can
detect object positions but requires that users explicitly specify how
to interact with each object.
\end{itemize}
\noindent By contrast, (4) \textit{SmartMove} integrates perception and planning via abstract queries for objects
matching a specified predicate. This condition allows users to more easily specify pick-and-place tasks via the addition of the new high-level \texttt{SmartGrasp} and \texttt{SmartRelease} operations.
Users found CoSTAR and Behavior Trees to be a usable and even enjoyable system; the higher levels of abstraction present in condition (4) allowed for improved task performance and better generalization to new environments.
\footnote[4]{Supplementary video of experiments and of the CoSTAR system in use is available on
YouTube, including expert instruction of the CoSTAR system and highlights from trials: {\scriptsize \url{https://www.youtube.com/watch?v=TPXcWU-5qfM\&list=PLF86ez-NVmyFMuj10dkUkgGlGpcM5Vok9}{https://www.youtube.com/watch?v=TPXcWU-5qfM\&list=PLF86ez-NVmyFMuj10dkUkgGlGpcM5Vok9}}}
In short, in this work we contribute:
\begin{itemize}
	\item Two new high-level actions to the existing CoSTAR framework, \texttt{SmartGrasp} and \texttt{SmartRelease}, designed to improve usability and generalization.
	\item Comparative analysis of different strategies for grounding end-user task plans via machine perception.
	\item Usability analysis of CoSTAR and each of its variants.
\end{itemize}

\section{Background}
Recent approaches for end-user instruction of collaborative robots include the development of new user interfaces~\cite{nguyen2013ros,mateo2014hammer,brunner2016rafcon}, learning from demonstration~\cite{ahmadzadeh2015learning,paxton2016want}, or systems that make use of natural language together with ontologies and large knowledge bases to follow high-level instructions, such as Tell Me Dave~\cite{misra2014tell} or
RoboSherlock~\cite{beetz2015robosherlock}.

Our proposed user interface is based on Behavior Trees, which have previously been used on humanoid and surgical robots, among other applications~\cite{bagnell2012integrated,marzinotto2014towards,hu2015semi}.
Others have explored designing user interfaces for robot task
specification~\cite{nguyen2013ros,brunner2016rafcon,paxton2017costar}. Nguyen et al.~\cite{nguyen2013ros} describe ROS Commander as a user interface based on finite state
machines for authoring task plans.
Similarly, Steinmetz and Weitschat~\cite{brunner2016rafcon} describe a
graphical tool called RAFCON.

Previous work in robot task specification has shown
complex~\cite{beetz2015robosherlock,toussaint2015logic} or interactive
behavior~\cite{dantam2011motion} without examining the specification of this behavior.
Dantam et al.~\cite{dantam2011motion} specify a complex motion grammar that allows a human to interactively play chess with a robot, where the robot must pick up and manipulate every piece on the board.
Methods for Task and Motion Planning allow planners to integrate selection of task parameterizations with computation of motion plans that will satisfy its requirements~\cite{toussaint2015logic,lagriffoul2014efficiently};
these methods are an inspiration for incorporating SmartMove into CoSTAR.

An alternate approach to direct task specification is to learn tasks from
expert demonstrations.
Alizadeh et al.~\cite{ahmadzadeh2015learning} learn skills which can be re-used according to a PDDL
planner.
Levine et al.~\cite{levine2015learning} proposed reinforcement learning methods for effectively learning individual skills with a demonstration as a prior.
In these cases, the end user still needs a way to connect individual skills.
Dianov et al.~\cite{dianov2016extracting} take a hybrid approach, using task graph learning to infer task structure from demonstrations and a detailed ontology.
Other recent work explored combining learned actions with sampling-based motion
planning and a high-level task specification~\cite{paxton2016want}.

\section{The CoSTAR System}\label{sec:system}
CoSTAR is a Behavior Tree-based user interface that aims to facilitate user interaction through a combination of an intuitive user interface, robust perception, and integrated planning and reasoning operations.
It is designed to be reliable, capable and cross-platform and was applied
to both the KUKA LBR iiwa and Universal Robots UR5 robot.
It is implemented as a component-based framework, where each component exposes a set of distinct operations that can be composed as a task plan~\cite{paxton2017costar}.

The underlying task is represented as a Behavior Tree (BT). BTs allow us to
visually construct complex, concurrent behaviors out of the equivalent of programming
constructs such as IF-statements, TRY-CATCH blocks, and FOR- and WHILE-loops.
In this section, we describe the basics of our BT implementation and the CoSTAR framework.
For more information on Behavior Trees, see prior work~\cite{guerin2015costar,paxton2017costar}.

\subsection{Overview of Behavior Trees}

\begin{figure*}[t]
\centering
\includegraphics[width=2\columnwidth]{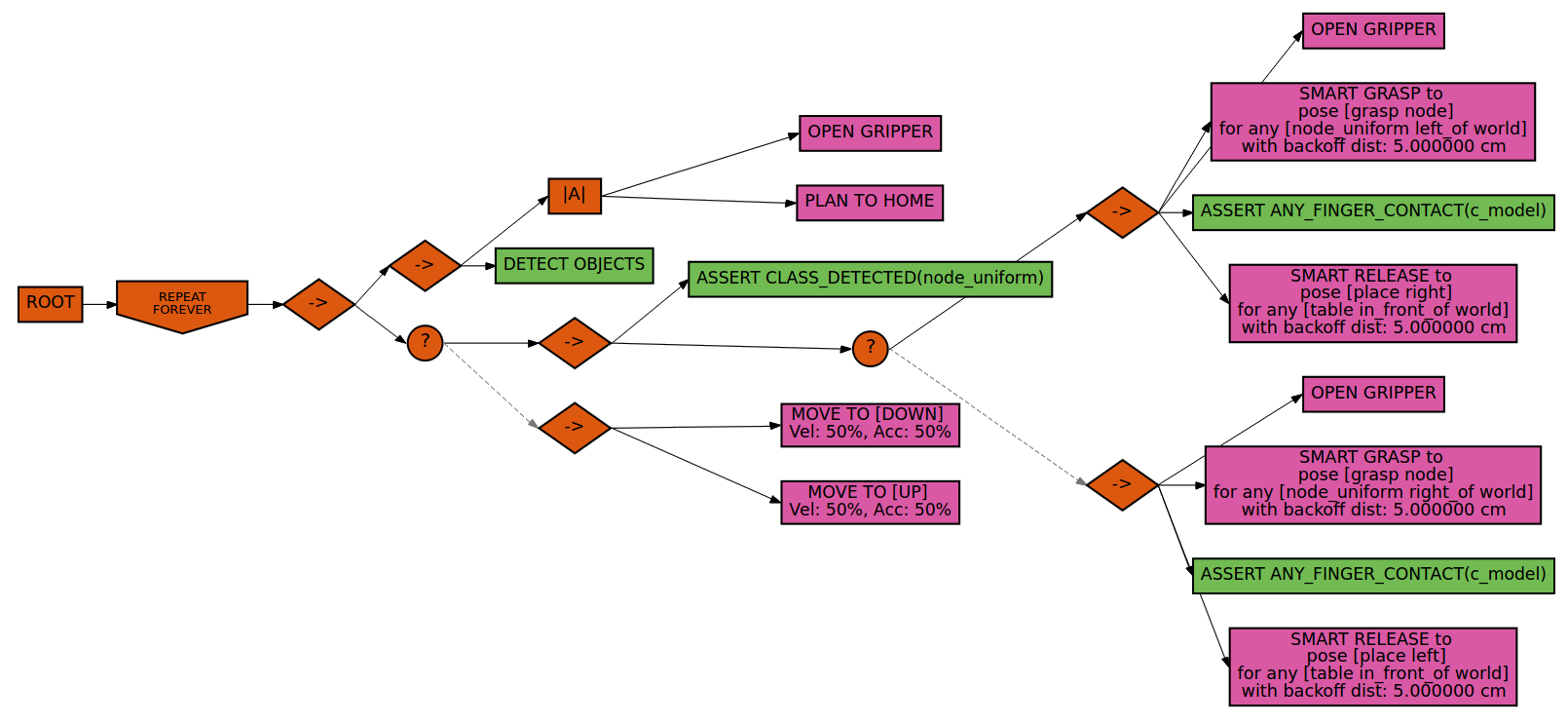}
\caption{A sample CoSTAR Behavior Tree displaying a complex task plan, including error checking and high-level queries to pick up and manipulate objects. Colors have been altered for readability.}
\label{fig:complex-task}
\end{figure*}

The basic operation of the BT is the ``tick:'' a status check sent at some high frequency (e.g., 60 hz) and propagated through the tree according to the rules associated with each node. Every node has an associated
action that is performed when it is ticked; if a node is ticked, it will return
some value in $\{\mathtt{SUCCESS, FAILURE, RUNNING}\}$.
The \textbf{Root} of a BT  has a single child node, generally a logical
node, and is the source of all ticks.

Complex task structure is achieved via \textbf{Logical} and decorator nodes. These control
program flow and the order of operations. CoSTAR's logical nodes include:
\begin{itemize}
  \item Sequence (\texttt{->}): Tick all children in order. This will stop when a
    tick reaches a running child.
  \item Selector (\texttt{?}): Tick all children in order until one returns
    success. This is used to construct IF-statements with multiple cases.
  \item Parallel-All (\texttt{|A|}): Tick all children in parallel; return
    success if all children return success and failure otherwise.
  \item Repeat: Tick all children some (possibly infinite) number of times
\end{itemize}
These elements can be combined to create relatively complex behavior.
Fig.~\ref{fig:complex-task} shows a tree that moves to the \texttt{Home}
joint position and opens the gripper in parallel, detects all objects in the
world, and then checks to see if it found any objects of class
\texttt{Node}.

\subsection{CoSTAR Architecture}\label{sec:architecture}

CoSTAR has a modular, component-based architecture.
What follows is a brief high-level overview of the CoSTAR system; for more
detail see Paxton et al.~\cite{paxton2017costar}.
In particular, we will refer to \textbf{Operations} $u$, which are specific actions that influence the world or update the robot's knowledge thereof. These are generally exposed as BT leaf nodes such as manipulation actions or calls to object detection software.
Operations can update the value of \textbf{Symbols} such as waypoints or known object positions, and also expose information as \textbf{Predicates} that make logical statements about the world like ``object A is to the left of object B.''

\begin{figure}[bt]
\centering
\includegraphics[width=0.5\textwidth]{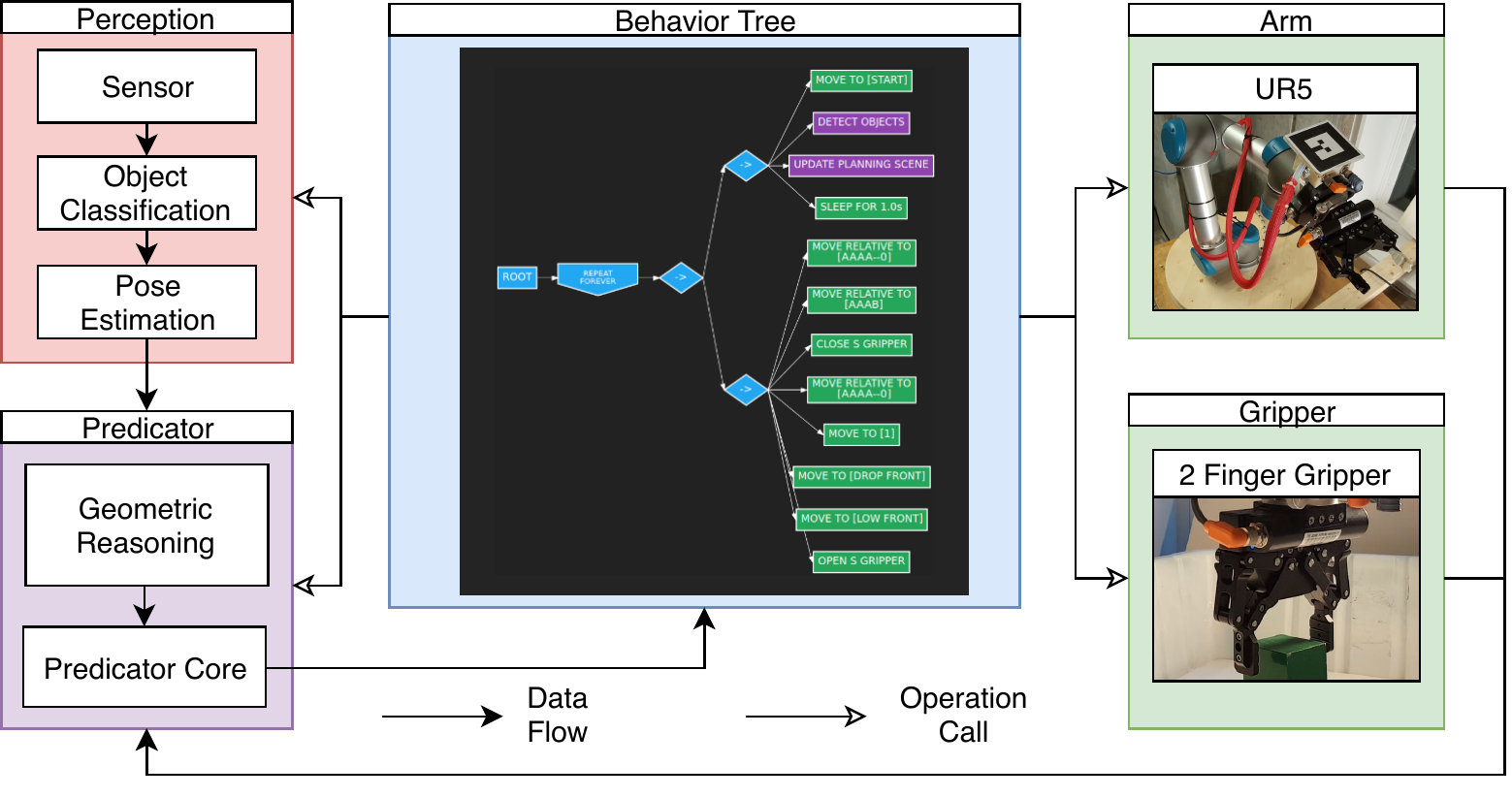}
\caption{A compressed overview of the CoSTAR system, based on the similar diagram in our previous work~\cite{paxton2017costar}.}
\label{fig:costar-system}
\vskip -0.5cm
\end{figure}

The main components of the CoSTAR system are shown in
Fig.~\ref{fig:costar-system}. In this work, we add the \texttt{SmartGrasp} and
\texttt{SmartRelease} operations, building on previous functionality, to allow
users to better perform pick and place tasks.

\subsubsection{Target Users}
Modern manufacturing is increasingly a high-tech field that requires strong problem-solving, math, and technical skills in its workers~\cite{giffi2015skills}.
As such, CoSTAR is designed for expert workers with a STEM background, though not necessarily with formal schooling.

\subsubsection{User Interface}
The heart of the CoSTAR UI is the BT-based task editor, which allows the end user to combine and
parameterize operations exposed to the user by all of these different
components.
Users can switch the robot into a compliant mode by pressing a
\texttt{TEACH} button, and can enable
autonomous execution by pressing the \texttt{SERVO} button.
Other functionality is hidden in a collapsible set of menus in the lower right,
which bring up dialogs for saving waypoints or SmartMove positions.
The task editor is accompanied by a 3D visualization of the robot, detected objects, and coordinate frames via the ROS RViz interface~\cite{quigley2009ros}.

\subsubsection{Perception and System Knowledge}

CoSTAR stores knowledge accumulated from a distributed set of sensors and components
through a special component called \texttt{Predicator}.
Predicator aggregates information about which objects have been detected,
what types of objects they are, and how they relate to each other and the task at hand.
It is exposed to end users through the \texttt{KnowledgeTest} and \texttt{PoseQuery} operations.
The \texttt{KnowledgeTest} operation checks to see if a certain predicate is
true. In Fig.~\ref{fig:complex-task}, we use knowledge tests to make sure that
a \texttt{node} object has been found, and to make sure that the  gripper successfully grasped an object.
One specific source of knowledge is the \texttt{DetectObjects} operation, which calls a 3D perception system to detect and estimate 6DOF poses for all objects in the scene~\cite{paxton2017costar}. For example, the \texttt{DetectObjects} operation appears near the top of the tree in Fig.~\ref{fig:complex-task}.
In this work, we also add a \texttt{DisableCollisions} operation, which allows us to specify that the robot can move closer than its minimum safe distance to a particular object (e.g. to pick it up or push it).

\subsubsection{Motion Planning and Execution}
The \texttt{Arm} component handles motion planning and execution, and ties in
closely with the \texttt{Predicator} component to expose more advanced
operations.
Grasping an object requires multiple steps: the robot must (a) align the gripper
properly with the grasp position, (b) move in, (c) close the gripper, and
finally (d) lift the object. The steps to this simple task plan remain the same
in a wide variety of tasks, being parameterized solely by the grasp frame and
the backoff distance. This backoff distance is used in steps (a) and (d) to
compute how far back from the object the robot will position its gripper. Placing an object is similarly complex.

In this work, we propose two unique actions to make pick and place tasks easier to specify in a general way: \texttt{SmartGrasp} and \texttt{SmartRelease}.
Each solves a planning problem for either grasping or 
placing with multiple possible goals based on user-provided information.
As a result, these two \texttt{SmartMoves} are more intuitive and reliable
than the capabilities found in previous versions of CoSTAR~\cite{paxton2017costar};
where operations had no memory of the specific object being manipulated and thus
sequential motions occasionally led to unexpected behavior.

\texttt{SmartGrasp} and \texttt{SmartRelease} query the \texttt{Predicator} component for objects
matching some set of conditions, and use object symmetry information to generate
a list of potential grasps. Then the \texttt{Arm} computes backoff poses and
sorts them based on joint-space distance. The resulting sorted list of grasp
poses is used to generate motion plans in order of preference via the
RRT-Connect algorithm~\cite{kuffner2000rrt} so that it will grab the closest
object that meets the given criteria.
In effect, users can then frame the task plan as a sequence of high-level commands such as:
{{
		\begin{align*}
		&\texttt{grasp(obj) with grasp\_position such that} \\
		&\texttt{is\_node(obj) and right\_of(robot, obj)}
		\end{align*}
}}
\noindent to grasp any \texttt{node} object on the right side of the robot.

\texttt{SmartRelease} is largely the same as \texttt{SmartGrasp}, but it
computes the backoff pose in the world frame's $-z$ axis and opens the gripper.
This is helpful when stacking objects or placing them on a table. The backoff
distance can be tuned by the user to achieve different behaviors. Again, the
\texttt{SmartRelease} is parameterized by a single demonstrated action and a
predicate specification.

\section{User Study}\label{sec:study}
\label{studydesign}

The goal of this study was to see how well users could use the operations
provided by each of four different perception strategies in order to adapt to an increasingly complex task.

Participants were randomly assigned to one of four groups, each associated with
a specific level of system capabilities and then asked to complete the study tasks. \ws{Fig.~\ref{fig:conditions} compares resulting task plans.}
This procedure was approved by the Johns Hopkins University Institutional Review Board (IRB) under protocol \#HIRB00005268.

\ws{
\begin{figure*}[t]
    \begin{multicols}{2}
        \centering
        \includegraphics[width=0.49\textwidth]{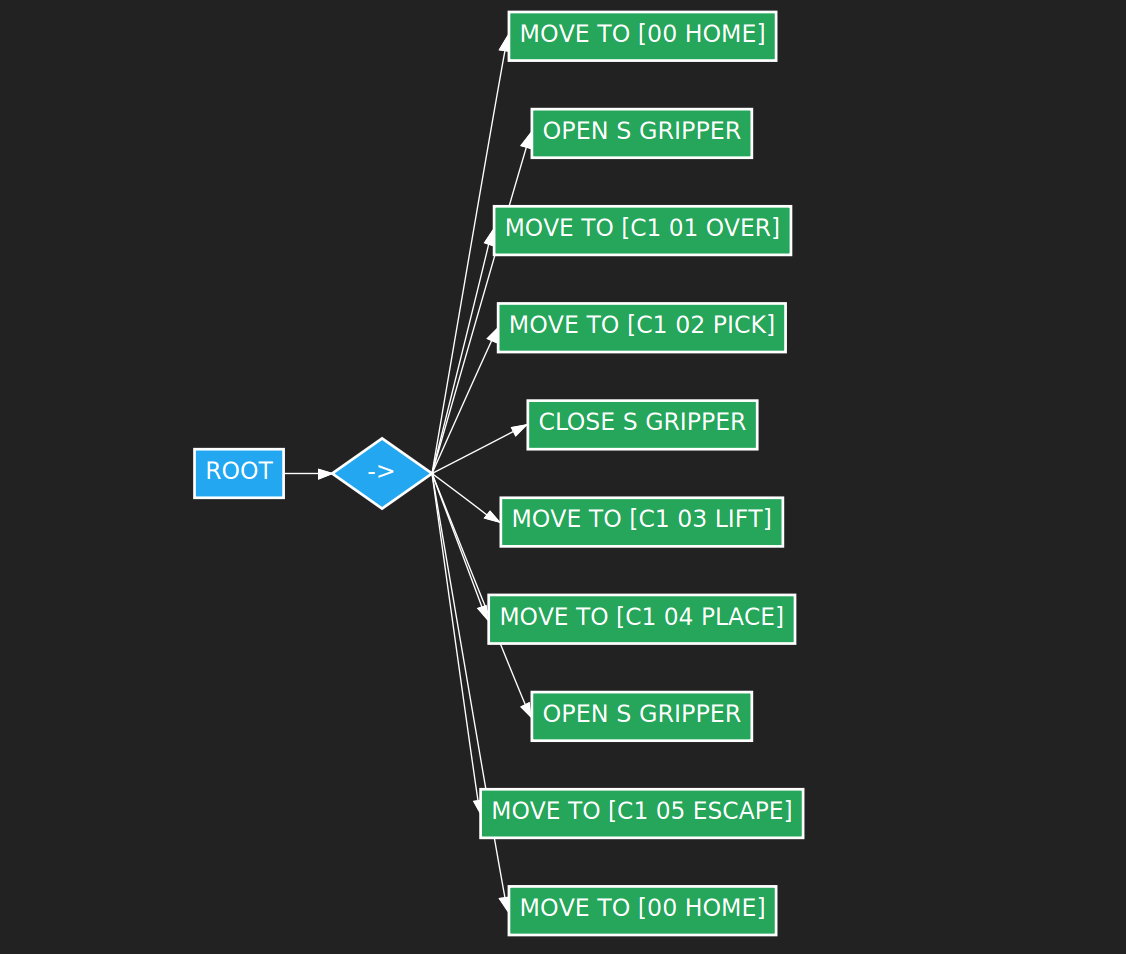}
        (1) Simple \\
        \includegraphics[width=0.49\textwidth]{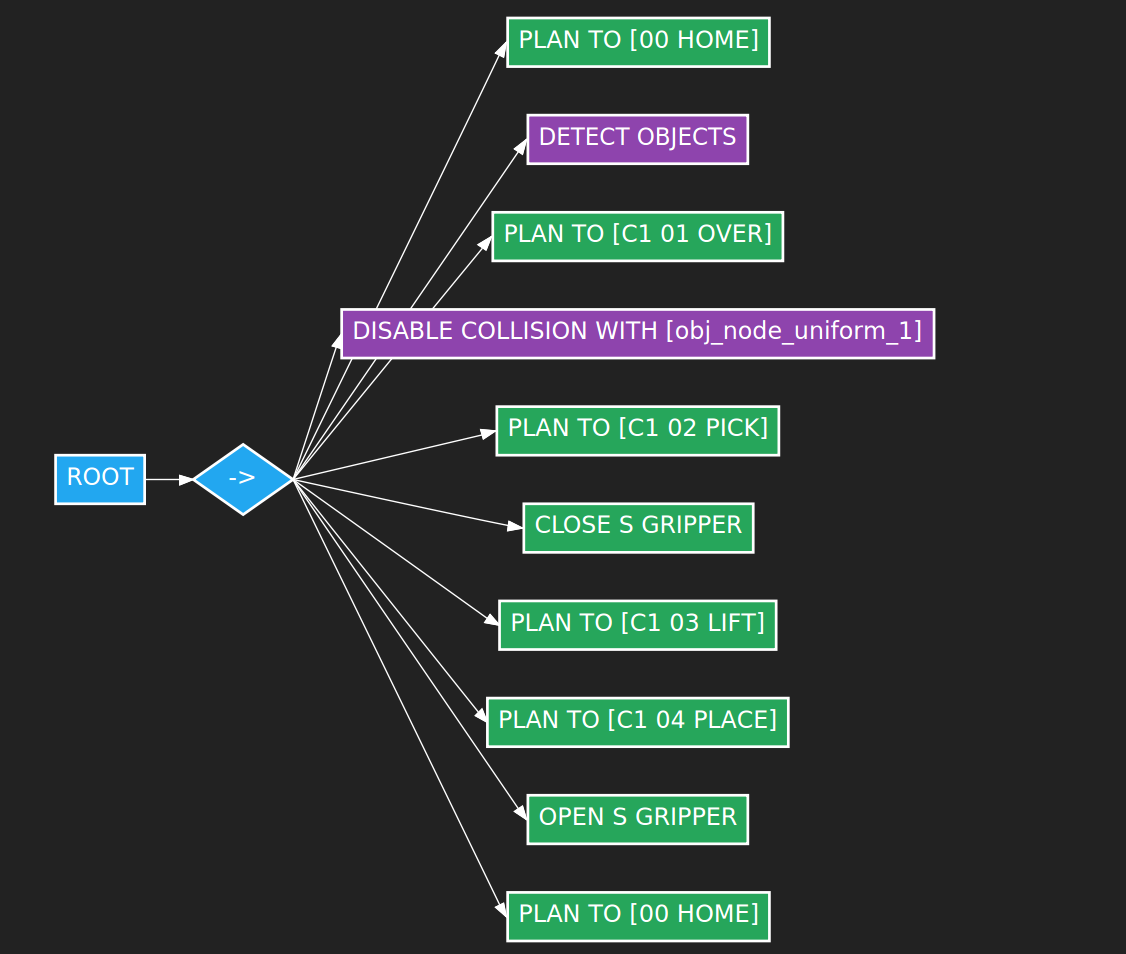}
        (2) Motion Planning \\
        \includegraphics[width=0.49\textwidth]{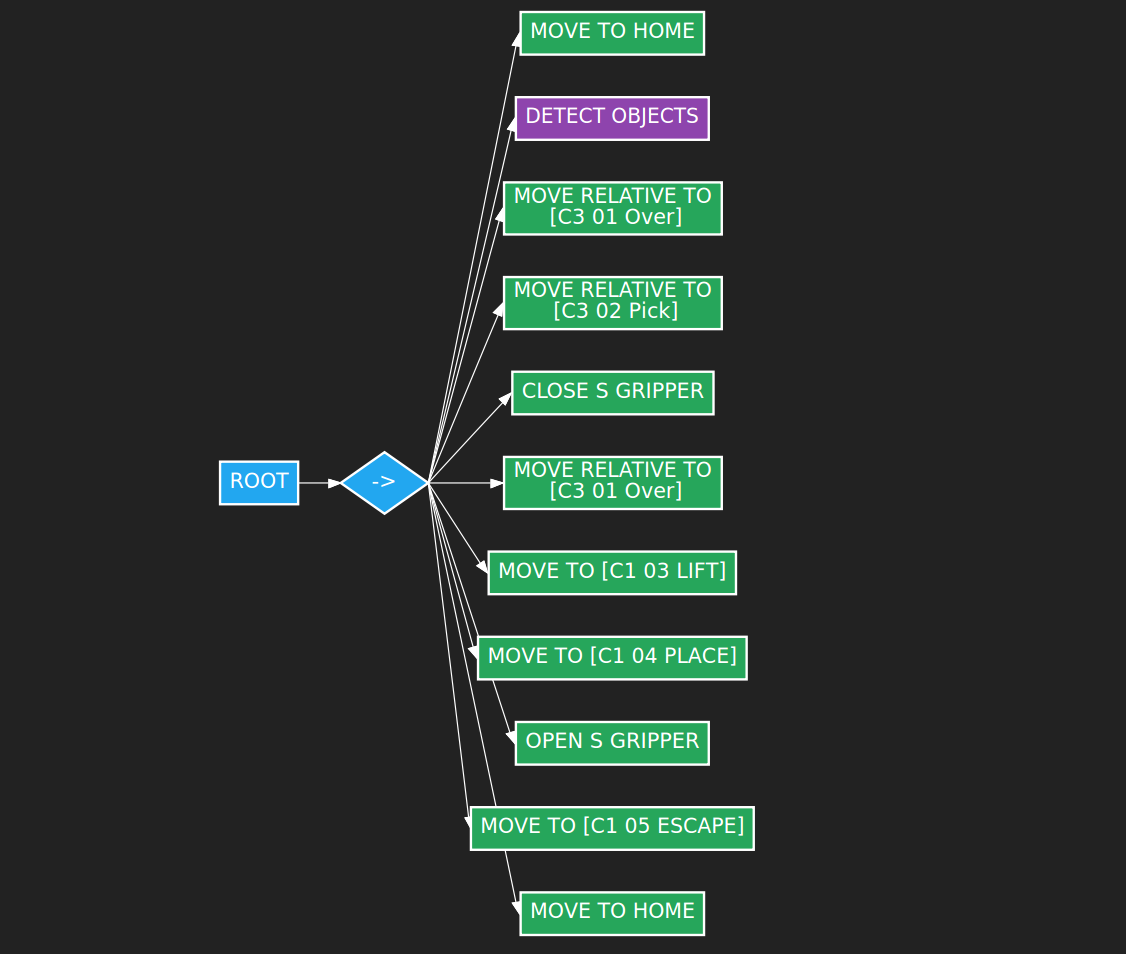}
        (3) Relative Motion \\
        \includegraphics[width=0.49\textwidth]{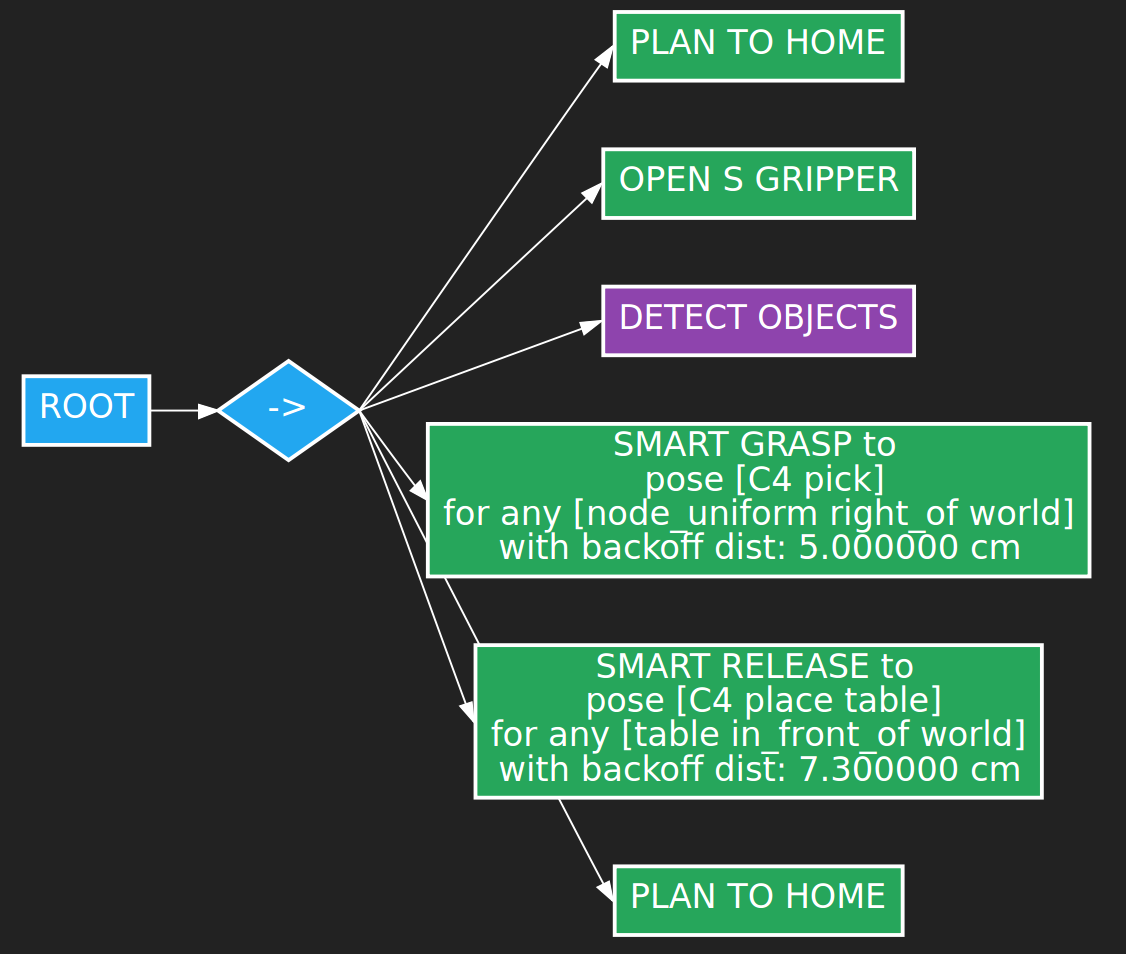}
        (4) SmartMove \\
    \end{multicols}
    \caption{Different CoSTAR operations were enabled and disabled under four different conditions to test different strategies for incorporating perception, as described in Section~\ref{studydesign}. All versions used the same BT-based user interface. Waypoints are expressed in square brackets (e.g. \texttt{C1 01 OVER}). Likewise, object names and parameters of \texttt{SmartMoves} appear in their respective boxes.}
    \label{fig:conditions}
    \vskip -0.5cm
\end{figure*}
}

\textbf{Condition 1 --- \emph{Simple}} represents a system similar to the commercially available programming environments of the Universal Robot UR5. The robot can open or close its gripper and move to positions relative to the robot base, but does not have access to any other CoSTAR capabilities including motion planning.

\textbf{Condition 2 --- \emph{Motion Planning}} adds motion planning: the user can update the
 scene model by detecting objects, but perception is only used to update scene geometry for the purposes of object avoidance. Thus, the robot will plan trajectories that avoid collisions and joint limits, but there is no ability to move relative to detected objects. The \texttt{DisableCollisions} operation allows the robot to move close to a specific object by disabling collision checking against it. This condition represents a ``naive'' approach of motion planning without abstraction.

\textbf{Condition 3 --- \emph{Relative Motion}} provides simple perception-based movements:
users have access to the \texttt{DetectObjects} operation and can define waypoints relative to detected object positions.
This represents the ``naive'' approach to incorporating object pose information where users must explicitly handle every object in the scene.

\textbf{Condition 4 --- \emph{SmartMove}} exposes two types of high-level actions: \texttt{SmartGrasp} and \texttt{SmartRelease}. They were asked to use these capabilities
and the \texttt{PlanToHome} action to complete the task.
These represent abstract, high-level
set of capabilities instead of the simpler, more explicit capabilities exposed in the other conditions.

\subsection{Study Tasks and Procedure}

We broke the study up into three phases: training, task performance, and the exit survey with subsequent generalization experiments on the user created trees.

\subsubsection*{Phase 1} We began with a short training phase, in which the participant moved a single block from the right to the left side of the robot. This task was not timed or used to score participants, and users could ask as many questions as they wanted.
Users were given an opportunity to ask any questions before moving on. This phase took 15 minutes.

\subsubsection*{Phase 2} Next we presented participants with three pick-and-place tasks with increasing
complexity. They were given 15 minutes to complete as much of each task as possible. 
These tasks were  designed to enable participants to incrementally learn and put into practice
how each UI component works, how the robot responds to user commands, and how
to build task plans using specific technologies such as perception and
planning.
All three tasks required that participants move square blocks called ``nodes'' in
different configurations from the right to the left of the robot without
knocking over an obstacle: a red ``link.''

\textbf{Task A --- \emph{Move Blocks}} asked participants to move two blocks from the right side of the workspace to the left.
Participants had to apply the knowledge from Phase 1 to teach the robot to move the second block themselves.
The obstacle was introduced to the world, but far enough away from the blocks that participants did not need to actively avoid it.

\begin{figure}[bt]
    \centering
    \subfloat[\label{fig:task2}]{
        \includegraphics[width=0.5\columnwidth]{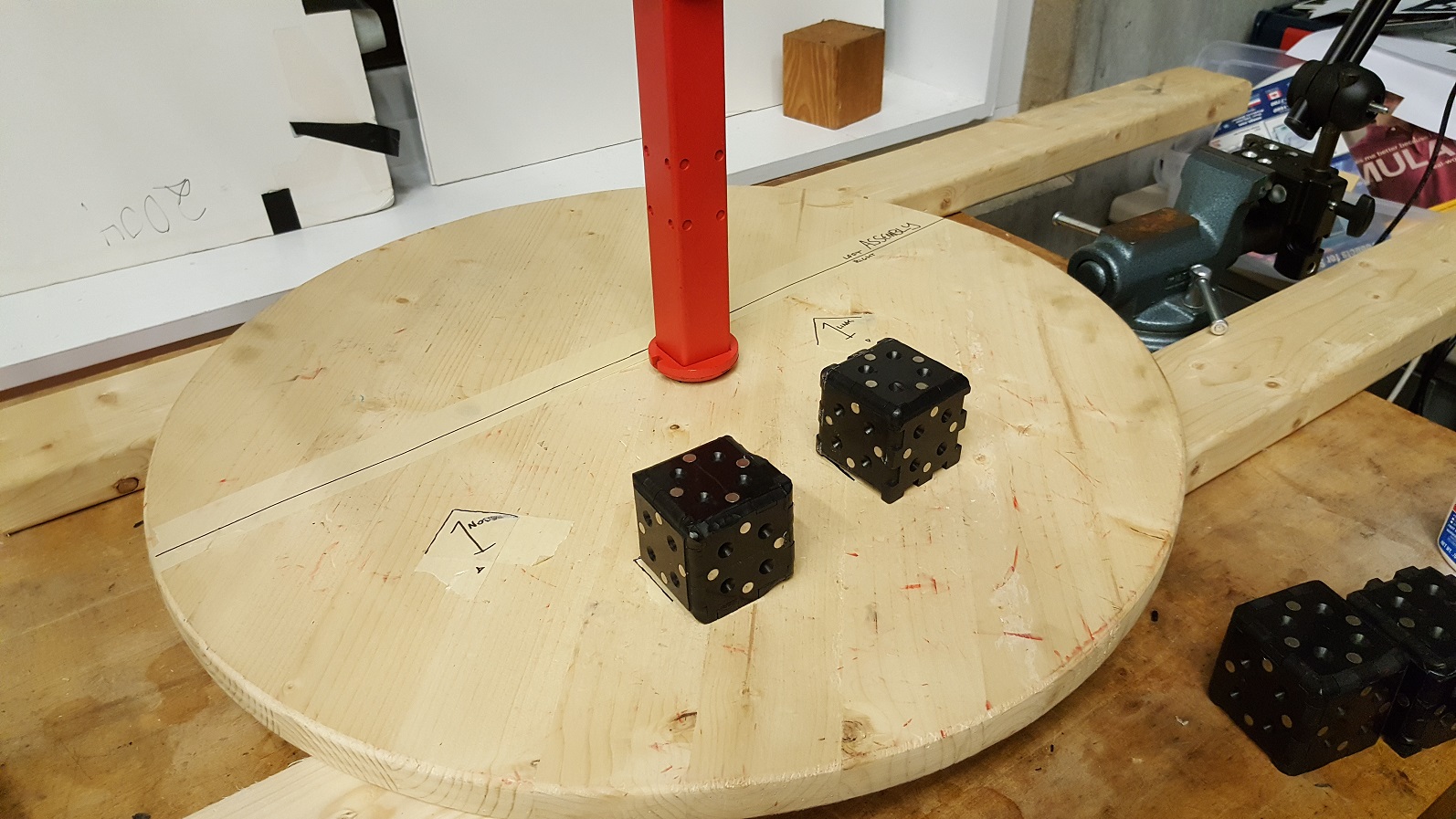}}
    \subfloat[\label{fig:task3}]{
        \includegraphics[width=0.5\columnwidth]{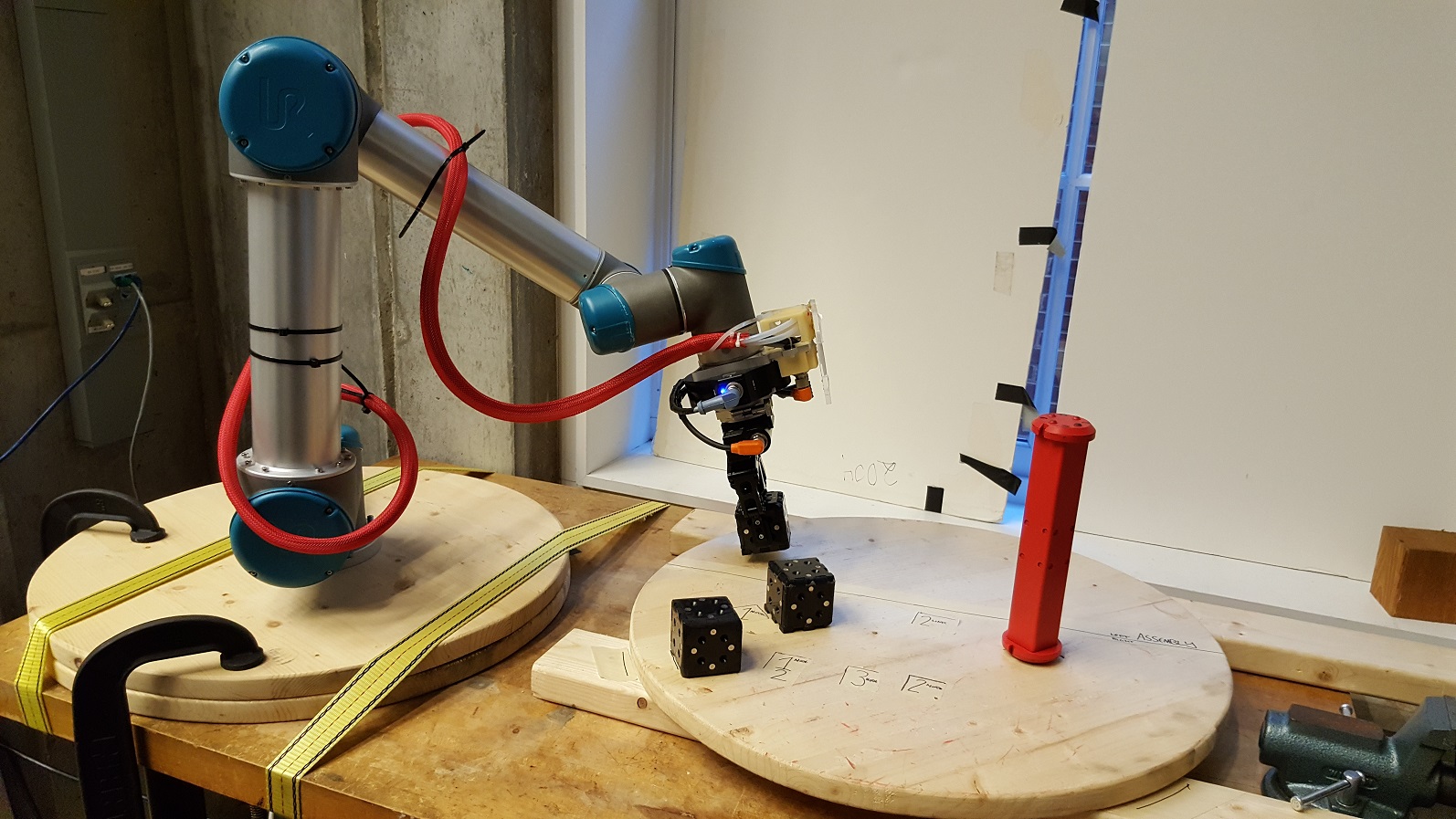}}
    \label{fig:tasks}
    \caption{Two of the study tasks. In Fig.~\ref{fig:task2}, users move two blocks from right to left; in Fig.~\ref{fig:task3} users can use any two blocks and additionally were asked to pick up the red link.}
\end{figure}

\textbf{Task B --- \emph{Avoid Obstacle}} required participants to similarly move two blocks
from the right to the left, although one of the blocks was placed in a
different position from the previous task and the obstacle was placed closer to
the two objects (Fig.~\ref{fig:task2}). 

\textbf{Task C --- \emph{Place Link}} presented participants with three blocks all of
which were in positions different from previous tasks. The link was moved
farther away again, and participants were asked to move two blocks of their
choice and to pick up the link and place it on top of one of these blocks. This
configuration is shown in Fig.~\ref{fig:task3}.

\subsubsection*{Phase 3} Finally, participants filled out a questionnaire that included the System Usability Scale \cite{albert2009beyond} and answered a set of interview questions.
After they left, we tested the generalization of user-authored task plans in two different cases
\ws{, shown in Fig.~\ref{fig:generalization}}.
In one case, we add additional obstacles to a scene and compare the motion planning vs. SmartMove condition. In the second case, we move objects to new positions to determine if SmartMoves can adapt better than Relative Motions to the new positions.

\ws{
\begin{figure}[bt]
\centering
\includegraphics[width=0.99\columnwidth]{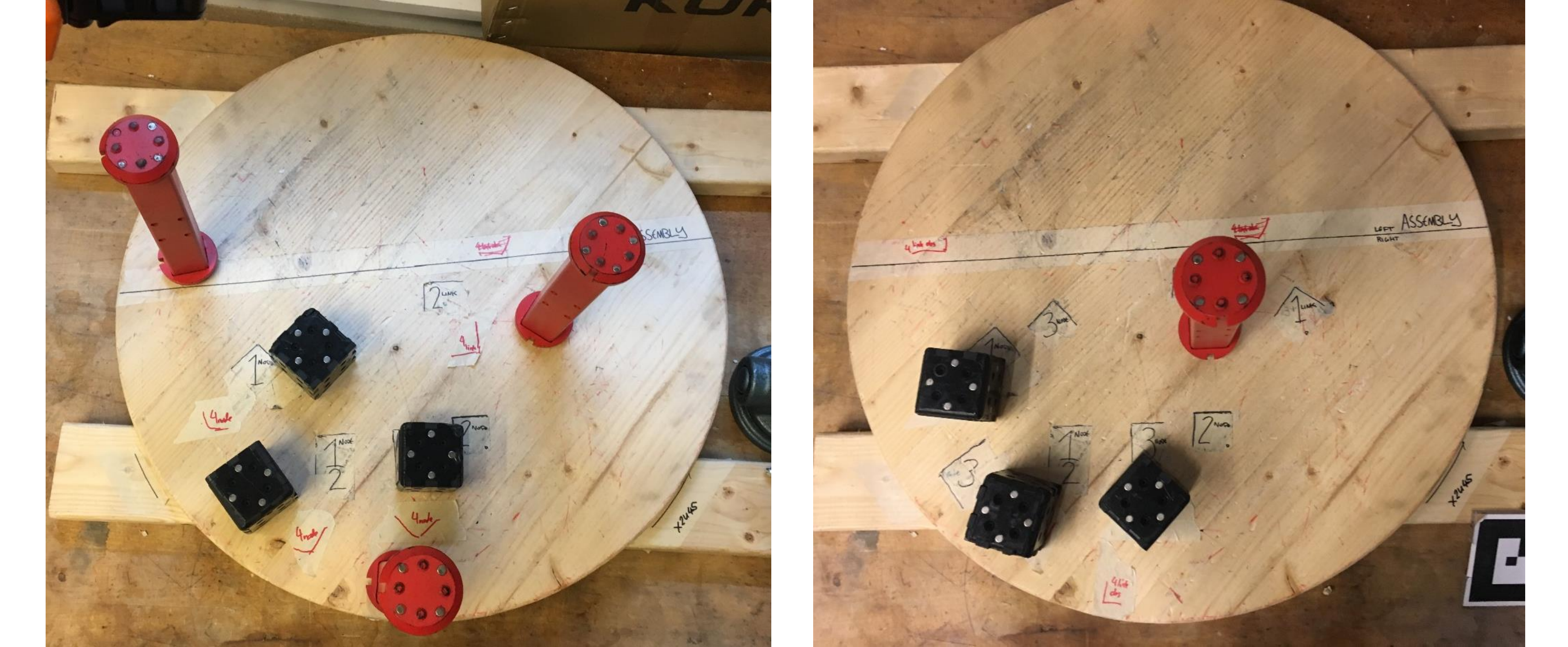}
\caption{Alternate world configurations. In the first case, additional links are
added to Task 3 as obstacles. In the second, all objects in Task 3 are moved to
new positions.}
\label{fig:generalization}
\vskip -0.5cm
\end{figure}
}

\subsection{Hypotheses and Metrics}

We examine three specific hypotheses in this study:
\begin{itemize}
\item[H1.] CoSTAR's SmartMove system will be perceived to be at least as usable as the baseline,
\item[H2.] When users are asked to generalize their Task 1 plan to a new
    environment configuration, users with the SmartMove condition
    will perform best.
\item[H3.] SmartMove plans will likewise perform best on generalization tasks
\end{itemize}

We developed a scoring metric to compare user performance. Users can attain a maximum of 1 point per component of the final assembly, with a bonus for fast task completion.
Performance on Task 1 and Task 2 was scored according to: $score = n_{nodes} -
n_{errors} + 2 \times (1-t)$,
where $t$ is the fraction of time in the trial used between 0 and 1, $n_{nodes}$ is the
number of nodes successfully moved from the right to the left, and $n_{errors}$
is 1 if the link was knocked over and 0 otherwise. This score was
configured to give a bonus to users who completed the task quickly, with a
strong score of $3$ if they used half of their time to finish the task.

We scored performance on Task 3 based on how many blocks were moved. The Behavior Trees
created by users were retained and tested after the end of the trial to compute
this score. The user received (1) one point for each of up to two nodes moved and
(2) one point if the link was successfully moved and placed. Due
to the difficulty of using a novel component, we give partial credit of 0.25
points for attempting to grasp the link, 0.75 points for moving and placing the
link but not achieving a successful mate, with 1.0 point reserved for a perfect
task performance. We give the same time-to-completion bonus as for Task 2 above.

We quantify perceived usability with a variant of the commonly used System
Usability Scale (SUS)~\cite{albert2009beyond}, which 
has been found to correlate well with other metrics for usability~\cite{bangor2009determining}. \ws{
Table~\ref{table:sus} shows the items included in the scale.

\begin{table}[bt]
  \centering
  \caption{Modified System Usability Scale for robotics user interface.
    Responses are scored from 1 (strongly disagree) to 5 (strongly agree).}
  \begin{tabular}{l p{7.3cm}} 
    \Xhline{2\arrayrulewidth}
    \# & Statement \\ 
    \hline
    1 & I think that I would use this interface frequently. \\
    2 & I found this interface unnecessarily complex. \\
    3 & I thought the interface was easy to use. \\
    4 & I would need the help of a robotics expert to be able to use this
    interface. \\
    5 & I found the various functions provided by this interface to be well
    integrated. \\
    6 & I thought the interface design was too inconsistent. \\
    7 & I imagine most people would learn to use this interface very quickly. \\
    8 & I found this interface cumbersome to use. \\
    9 & I feel very confident in using this interface. \\
    10 & I need to learn a lot of things before I could be an effective user of this
    interface. \\
    \Xhline{2\arrayrulewidth}
  \end{tabular}
  \label{table:sus}
\end{table}
}

\subsection{Participants}
Our target demographic represents highly skilled manufacturing workers with a STEM background~\cite{giffi2015skills}.
To approximate this demographic, we performed our study on undergraduate and graduate students in math, science,
engineering, and computer science:
users who are technically savvy with some education but who are not necessarily experts in robotics.
Subjects were recruited via email.

We recruited at least 8 users for each of 4 conditions, in line with accepted practices for studies in this category\cite{caine2016local}.
Altogether, $40$ users participated in the study.
Five users were excluded for the following reasons: three were excluded for being non-technical novice users and not undergraduate or
graduate students, and two users were excluded due to issues with the perception system during their trials.
Characteristics of the population are described in
Table~\ref{table:participants}.
Our user pool is less familiar with robotics ($M=3.5$, $SD=1.8$) than with programming ($M=5.6$, $SD = 1.6$). Not coincidentally, many of our test subjects
were computer science students.

\subsection{Study Results}\label{sec:results}

Data from the System Usability Scale~\cite{albert2009beyond} indicate that users found CoSTAR to be highly usable with an average SUS score of $73.4$ out of $100$ across all users.
Previous work has found that a system with ``good'' usability will have a mean
score of $71.4$, while a system with ``excellent'' usability will have a mean score
of 85.5~\cite{bangor2009determining}.
In aggregate, they considered CoSTAR to be good, but not a perfectly usable system.

\begin{table}[bt] 
\centering
\caption{Self-reported participant population characteristics.}
\begin{tabularx}{\columnwidth}{l l l}
\Xhline{2\arrayrulewidth}
\multicolumn{2}{l}{Population Characteristic} & Value \\
\hline
\multicolumn{2}{l}{Average Age} & $23.0 \pm 3.1$ \\
\multirow{2}{*}{Gender} & \# Male & 25 \\
& \# Female & 10 \\
\multirow{3}{*}{Familiarity (1 to 7)}& Robotics & $3.5 \pm 1.8$ \\
& Programming & $5.6  \pm 1.6$ \\
& Video Games & $3.8 \pm 1.8$ \\
\multirow{5}{*}{Major} & Computer Science & 15 \\
& Robotics & 4 \\
& Other Engineering & 7 \\
& Other Math/Science & 2 \\
& Unspecified & 7 \\
\Xhline{2\arrayrulewidth}
\end{tabularx}
\label{table:participants}
\end{table}

\begin{figure}[bt]
  \centering
  \includegraphics[width=\columnwidth]{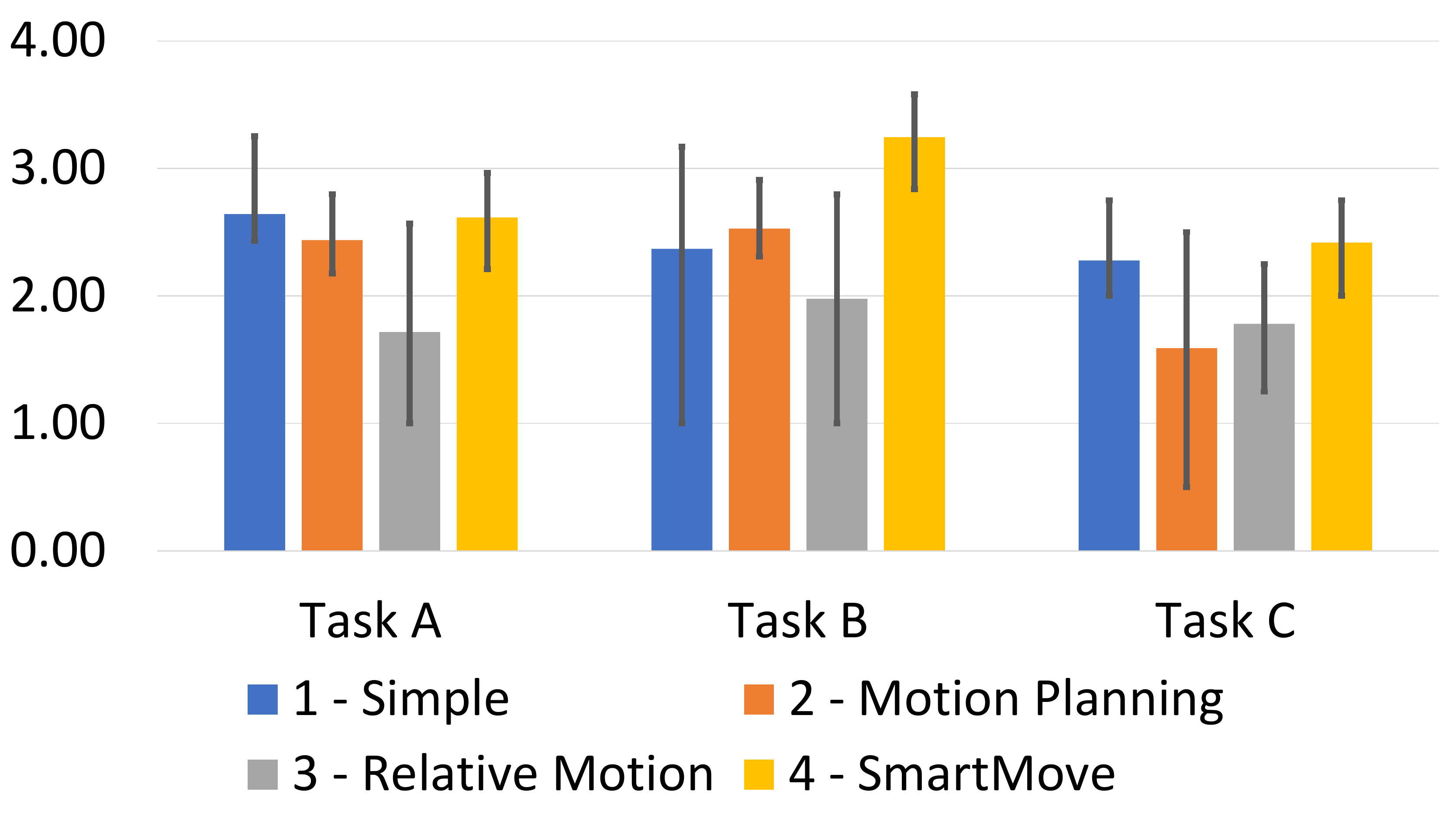}
  \caption{Scores for each condition on tasks A-C. Error bars indicate interquartile range. Results showed
  a wide variance of final scores for Task C in particular; users had trouble understanding how best to use many variants of the system. }
  \label{fig:scores}
  \vskip -0.5cm
\end{figure}

Table~\ref{table:task-results} summarizes the task performance and usability
data collected in the study.
We limited the time users were able to spend on each task in order to see some variability in task success rates, since all versions of the system were capable of completing all tasks given enough time.
In general, users performed similarly on Task A, as seen in Fig.~\ref{fig:scores}.

\begin{table*}[bt]
\centering
\caption{Results from each of the four conditions on the three tasks showing
performance times and self-reported System Usability Scale (SUS) scores after
completion of the experiment.}
\centering
\begin{tabular}{ l l | l l l | l l l | l l l }
\Xhline{2\arrayrulewidth}
 & & \multicolumn{3}{c}{\centering {Task A}} & \multicolumn{3}{c}{\centering {Task B}} & \multicolumn{2}{c}{\centering {Task C}} & Total \\
Condition & \# Users & Score & \% Success & Time & Score & \% Success
& Time & Score & SUS & Score\\
\hline
1 Simple & 9 & $2.64 \pm 0.73$ & 88.9\% & 8:39 & $2.37 \pm 1.04$ & 66.7\% & 7:06 & $2.28 \pm 0.52$ & $73.8 \pm 8.7$ & $7.15 \pm 1.78$ \\

2 Motion Planning & 9 & $2.44 \pm 0.32$ & 100.0\% & 11:43 & $2.53 \pm 0.66$ & 88.9\% & 9:36 & $1.59 \pm 1.13$ & 74.2 $\pm 10.1$ & $6.58 \pm 1.67$ \\

3 Relative Motion & 8 & $1.72 \pm 0.97$ & 71.4\% & 12:24 & $1.98 \pm 1.13$ & 62.5\% & 9:16 & $2.1 \pm 1.24$ & $69.5 \pm 10.4$ & $5.56 \pm 1.96$ \\

4 SmartMove & 9 & $2.62 \pm 0.35$ & 100.0\% & 10:23 & $3.24 \pm 0.47$ & 100.0\% & 5:40 & $2.42 \pm 0.50$ & $75.3 \pm 6.7$ & $8.31 \pm 0.34$ \\
\Xhline{2\arrayrulewidth}
\end{tabular}
\label{table:task-results}
\vskip -0.5cm
\end{table*}

\subsubsection*{Task B}
SmartMove users performed significantly better than Motion Planning users
on Task B ($p=0.0096$) or
Relative Motion users ($p=0.0079$) according to a pairwise T-test on our task completion metric. They only performed marginally better than
users of the Simple condition ($p=0.0207$). All significance results use Bonferroni-corrected alpha levels of $0.05 / 3 = 0.0167$.

\subsubsection*{Task C}
User performance varied the most on Task C, most likely due to the difficulty of learning how to manipulate a new object.
Scores from SmartMove are still higher than those from the Motion Planning and Relative Motion
groups ($p=0.035$ and $p=0.048$, respectively).
Differences are even clearer when comparing total scores across all three tasks as shown in Table~\ref{table:task-results}: SmartMove is better than both Motion Planning ($p=0.007$) and Relative Motions ($p=0.003$).

\subsubsection*{Generalization}
The SmartMove system also had a higher score than Motion Planning at the generalization task with extra obstacles ($p=0.029$). SmartMove scored an average of $1.65 \pm 0.89$, Motion Planning scored an average of $0.43 \pm 1.28$.
On the generalization task with changed object positions, it likewise scored higher than the Relative Motion test case ($p=0.056$), with an average of $2.25 \pm 0.35$ vs. $1.5 \pm 1.29$. We saw much more variation on the object pose generalization task for the Relative Motions task: some users' trees generalized very well, others very poorly. SmartMove task plans were comparatively very consistent across users.

\section{Discussion}\label{sec:discussion}
There are three key findings from this study.
First, these results show that users find Behavior Trees to be a
practical and effective means of defining a robot program.
Second, perceptual abstractions such as SmartMove allow end users to more easily specify and adapt robot programs.
Finally, such programs are more general and more robust to environmental variation.
Users can employ advanced robotic capabilities if exposed to them properly and with proper training.
For motion planning and perception to make a substantial impact they must support the user's mental model of the task.

Users were comfortable with the mixture of hands-on teaching and editing the Behavior Tree, as shown in Fig.~\ref{fig:users}.
Participants assigned to the Simple or Motion Planning groups found the robot
easy to manage and predictable,
but they expressed frustration by the degree to which they
had to micro-manage positioning the robot by specifying multiple waypoints.
One user operating under the Simple condition complained that in
a large tree, ``replacing them one by one every time is a little inconvenient.''
The effect was clear when users were asked to generalize their plan for Task 1
to solve Task 2: SmartMove users clearly outperformed the others, lending strong support to H1. Indeed, some users of the Simple condition struggled with this generalization because they had so many waypoints that needed to change.

We found evidence for H2, noting that our perceptually abstracted SmartMoves offered superior generalization to Motion Planning or Relative Motion alone. When tested on more challenging scenarios with additional obstacles, trees using SmartMove were able to adapt by choosing an alternate grasp or an alternate object to pick up when a more explicit method for specifying the task would have failed. The Simple condition, while easy to use, had no ability to generalize whatsoever.

Perception was not useful to our participants if they could not communicate effectively with the robot.
Compare the Simple condition with Motion Planning and Relative Motion in Fig.~\ref{fig:scores}. It offers no built-in problem solving ability (resulting in high variance in Task B) but is comparable or better than these conditions. 
Users of Motion Planning and Relative Motion often could not predict what the robot would do or why it would do it. As a result, these were less effective or consistent than either the SmartMove or Simple conditions.
Participants were often unclear on when knowledge they provided to the robot would generalize and when it would not.
One user assigned to the Relative Motion condition felt the robot ``spazzed out,'' i.e., it did not do what they were expecting.

SmartMove was more predictable, although there is room to improve the way the design and implementation details are communicated within the interface.
For instance, users often believed that \texttt{SmartGrasp}
operations had to be re-taught for every node object, but this is not necessary.
On the other hand, it was also not clear that new drop positions must be taught to
\texttt{SmartRelease} for every new node in the workspace.
The physical meaning of the arguments for the SmartMove query could also benefit from improved clarity. For example,
One user said they ``were not able to figure out why [the next block] was left,'' they only knew that it worked. These points suggest areas for future improvement.

\section{Conclusions}\label{sec:conclusions}

In conclusion, our results show that abstract perception allows users to construct robust, generalizable task plans without negatively impacting usability. 
In addition, behavior Trees can form the basis for a powerful, flexible, and highly usable visual programming language for collaborative robots.

\section*{ACKNOWLEDGMENT}
This work was funded by NSF Award No. 1637949.

\bibliographystyle{IEEEtran}
\bibliography{kuka,lfd,planners,software,vision,costar,psych,task_description}

\end{document}